\documentclass{article} % For LaTeX2e
\usepackage{longcat_style,times}

\usepackage{amsmath,amsfonts,bm}

\def\eqref#1{equation~\ref{#1}}
\def\1{\bm{1}}

\DeclareMathAlphabet{\mathsfit}{\encodingdefault}{\sfdefault}{m}{sl}
\SetMathAlphabet{\mathsfit}{bold}{\encodingdefault}{\sfdefault}{bx}{n}

\usepackage{hyperref}
\usepackage{url}

\usepackage{graphicx}
\usepackage{balance}  % for  \balance command ON LAST PAGE  (only there!)
\usepackage{amsmath}
\usepackage{algpseudocode}
\usepackage{latexsym}
\usepackage{multirow}
\usepackage{multicol}
\usepackage{color}
\usepackage{dashrule}
\usepackage{extarrows}
\usepackage{dsfont}
\usepackage{centernot}
\usepackage{hyperref}
\usepackage{natbib}
\usepackage{fancybox}
\usepackage[utf8]{inputenc} % allow utf-8 input
\usepackage[T1]{fontenc}    % use 8-bit T1 fonts
\usepackage{hyperref}       % hyperlinks
\usepackage{url}            % simple URL typesetting
\usepackage{booktabs}       % professional-quality tables
\usepackage{amsfonts}       % blackboard math symbols
\usepackage{amsmath}
\usepackage{nicefrac}       % compact symbols for 1/2, etc.
\usepackage{microtype}      % microtypography
\usepackage{xcolor} % colors
\usepackage{amsmath}
\usepackage{lipsum}
\usepackage{colortbl}
\usepackage{subcaption}
\usepackage{tabularx}
\usepackage{makecell}
\usepackage{wrapfig}
\usepackage{bm}
\usepackage{multirow}
\usepackage{xcolor} % Load the color package
\usepackage{bbm}
\usepackage{mdframed}
\usepackage{enumitem}
\usepackage{xspace}

\usepackage{lineno}

\definecolor{darkblue}{rgb}{0, 0, 0.5}
\hypersetup{colorlinks=true, citecolor=darkblue, linkcolor=darkblue, urlcolor=darkblue}

\usepackage{listings}
\usepackage{xcolor}
\usepackage[ruled,vlined]{algorithm2e}
\usepackage{caption}
\usepackage{float} % for H option in algorithm
\usepackage{arydshln}
\usepackage{wrapfig} 
\usepackage{tcolorbox}
\tcbuselibrary{theorems}
\tcbuselibrary{most}
\usepackage[dvipsnames]{xcolor}

\usepackage{hyperref}       % hyperlinks
\usepackage{url}            % simple URL typesetting
\usepackage{booktabs}       % professional-quality tables
\usepackage{amsfonts}       % blackboard math symbols
\usepackage{nicefrac}       % compact symbols for 1/2, etc.
\usepackage{microtype}      % microtypography
\usepackage{xcolor}         % colors
\usepackage{amsmath}
\usepackage{multirow}
\usepackage{multicol}
\usepackage{colortbl}
\usepackage{tcolorbox}
\usepackage{wrapfig}

\usepackage{cleveref}
\usepackage{xspace}

\makeatletter
\DeclareRobustCommand\onedot{\futurelet\@let@token\@onedot}
\def\@onedot{\ifx\@let@token.\else.\null\fi\xspace}

\def\eg{\emph{e.g}\onedot} 
\def\ie{\emph{i.e}\onedot} 
 
\def\etc{\emph{etc}\onedot}

\makeatother

\usepackage{natbib}

\definecolor{light-gray}{gray}{0.6}
\definecolor{front-color}{HTML}{F5FFFA}
\definecolor{Gray}{gray}{0.93}

\definecolor{customTeal}{RGB}{0, 128, 128} 
\definecolor{emphasisColor}{RGB}{255, 0, 0} % Red color for emphasis
\definecolor{oursBlue}{RGB}{51,202,246}

\definecolor{blue1}{HTML}{508AB2}
\definecolor{green2}{HTML}{BFF6BA}

\newtcbtheorem[]{prompt}{Prompt}%
{colback=SeaGreen!10!CornflowerBlue!10,
 colframe=RoyalPurple!55!Aquamarine!100!,
 fonttitle=\bfseries,
 left=.02in, right=.02in, bottom=.02in, top=.02in,
 before upper={\linespread{1.5}\selectfont}}{prompt}

\renewcommand{\shorttitle}{AIA: Rethinking Architecture Decoupling Strategy In Unified Multimodal Model}
\definecolor{darkblue}{rgb}{0, 0, 0.5}
\hypersetup{colorlinks=true, citecolor=darkblue, linkcolor=darkblue, urlcolor=darkblue}

\makeatletter
\renewcommand{\@maketitle}{%
  \vbox{%
    \hsize\textwidth
    \linewidth\hsize
    \vskip -0.5in
    \noindent
    \begin{minipage}{0.99\textwidth}
  \includegraphics[width=0.27\linewidth]{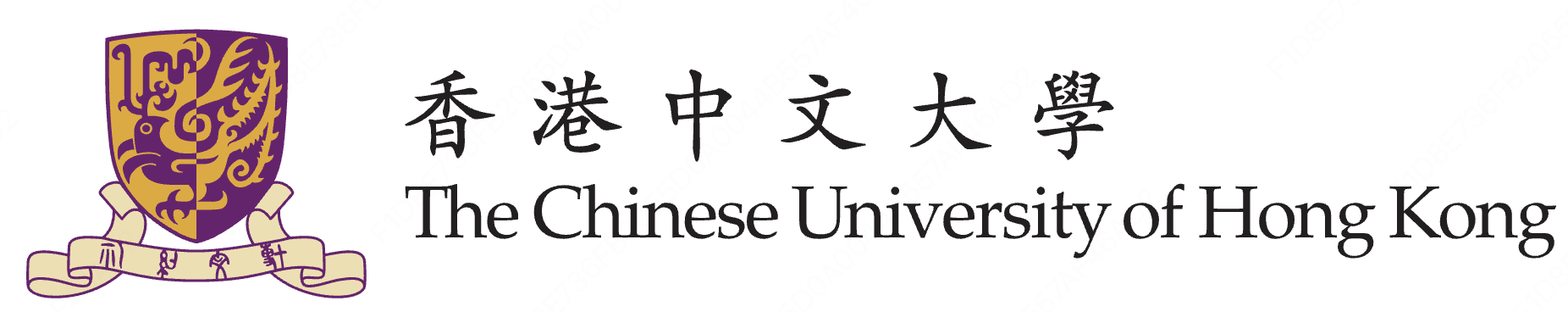}
    \end{minipage}%
    \\
    \rule{\linewidth}{1pt}
    \hspace{0.05\textwidth}%
    \begin{minipage}{0.8\textwidth}
    \end{minipage}

    \centering
    {\LARGE \bfseries\@title\par}
    \vskip 0.1in  % 调整这个值：0.3in=小, 0.5in=中, 0.7in=大
    \def\And{%
      \end{tabular}\hfil\linebreak[0]\hfil%
      \begin{tabular}[t]{c}\bf\rule{\z@}{24\p@}\ignorespaces%
    }
    \def\AND{%
      \end{tabular}\hfil\linebreak[4]\hfil%
      \begin{tabular}[t]{c}\bf\rule{\z@}{24\p@}\ignorespaces%
    }
    \begin{tabular}[t]{c}\bf\rule{\z@}{24\p@}\@author\end{tabular}%
  \vskip 0.05in 
  }
}
\makeatother

\title{AIA: Rethinking Architecture Decoupling Strategy In Unified Multimodal Model} 

\makeatletter
\def\@fnsymbol#1{\ensuremath{\ifcase#1\or \dagger\or \ddagger\or
   \mathsection\or \mathparagraph\or \|\or **\or \dagger\dagger
   \or \ddagger\ddagger \else\@ctrerr\fi}}
\makeatother

\newcommand{\homepage}{\raisebox{-1.5pt}{\includegraphics[height=1em]{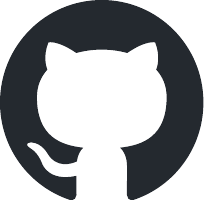}}}
\newcommand{\hfmodel}{\raisebox{-1.5pt}{\includegraphics[height=1em]{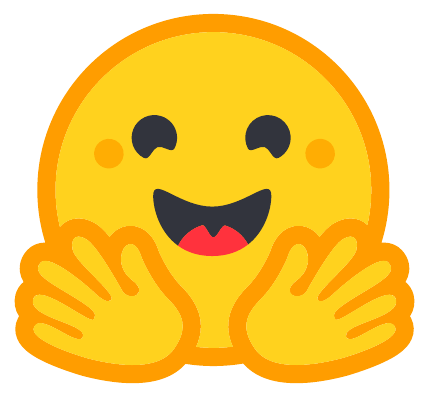}}}

\author{
\begin{tabular}{c}
\textbf{Dian Zheng}$^{1}$
\quad
\textbf{Manyuan Zhang}$^{1,2}$\thanks{Project Leader.}
\quad
\textbf{Hongyu Li}$^{2}$
\quad
\textbf{Kai Zou}$^{3}$
\quad
\textbf{Hongbo Liu}$^{4}$ \\[1ex]
\textbf{Ziyu Guo}$^{2}$
\quad
\textbf{Kaituo Feng}$^{1}$
\quad
\textbf{Yexin Liu}$^{2}$
\quad
\textbf{Ying Luo}$^{2}$
\quad
\textbf{Hongsheng Li}$^{1}$\thanks{Corresponding Author.} \\[1ex]
\normalfont $^1$MMLab, CUHK 
\quad
$^2$Meituan
\quad
$^3$USTC
\quad
$^4$TJU\\[1ex]
{\homepage\ \normalfont 
\texttt{Home: \!\!\!\!\!\url{https://github.com/zhengdian1/AIA}}} \\
{\hfmodel\ \normalfont \texttt{HF: \!\!\!\url{https://huggingface.co/zhengli1013/AIA}}} \\
\end{tabular}
}

\begin{document}
\maketitle

% {%
%    \renewcommand\twocolumn[1][]{#1}%
%    \maketitle
%    \vspace{-1pt}
%    \begin{center}
%     \centering
%     \includegraphics[width=0.99\linewidth]{figures/teaser.pdf
%     }
%     \captionof{figure}{.}
%     \vspace{6pt}
%     \label{fig:teaser}
%    \end{center}%
%   }
   
\begin{abstract}
  Unified multimodal models for image generation and understanding represent a significant step toward AGI and have attracted widespread attention from researchers. The main challenge of this task lies in the difficulty in establishing an optimal training paradigm due to inherent conflicting targets in understanding and generation tasks. To alleviate these conflicts and pursue higher performance, many researchers adopt varying degrees of architecture decoupling (\eg, Double image encoders, MOE/MOT architecture, or frozen MLLM). However, excessive model decoupling can lead to the loss of interleave generation ability, undermining the original intent of unified models. In this work, we aim to explore how to mitigate task conflicts without resorting to model decoupling. Firstly, we analyze why decoupling boosts performance by studying the cross-modal attention behavior of models. We observe that \textbf{architecture decoupling does not solve task conflicts}, but essentially drives models toward cross-modal interaction patterns of task-specific models, as seen in Qwen3-VL and HunyuanImage-3.0, and that the more thorough the decoupling, the more consistent the behavior becomes. Motivated by this observation, we propose \textbf{A}ttention \textbf{I}nteraction \textbf{A}lignment (\textbf{AIA}) loss, which explicitly learns task-specific multimodal interaction patterns during training. To demonstrate the generalizability of our AIA loss, we apply it to Emu3 and Janus-Pro during SFT and post-training stage respectively. Without bells and whistles, AIA not only refines cross-modal attention patterns, but also boosts both generation and understanding performance.
\end{abstract}

\section{Introduction}
\label{sec:intro}
Unified Multimodal Model (UMM) is trained to perform two distinct tasks (\eg, visual generation and understanding) within a single network, aiming to enhance interpretability by visualizing intermediate processes (interleaved generation), while also improving single-task performance. This approach represents a significant step toward general artificial intelligence.

Although the original intention behind UMM is admirable, practical realities are harsh: visual understanding and generation tasks require distinct feature granularities and representations at different network layers. Early research~\cite{wang2024emu3,wu2024liquid,team2024chameleon} attempted fully unified architectures (\eg, sharing the image encoder and base model), but the results lagged significantly behind single-task approaches. To address task conflicts and boost performance, some researches~\cite{januspro,wu2025omnigen2,li2025onecat,bagel} have begun to decouple model components to varying degrees, achieving promising results. Due to the inherent effectiveness of this strategy, more researchers are now pursuing decoupled architectures.
However, this trend overlooks the core motivation of UMM~\cite{unimmmu,yang2025survey}: \textit{leveraging the capacity of the unified model for cross-modal reasoning to enhance single-task performance.} Excessive decoupling risks losing this synergistic benefit, limiting the model's ability to transfer knowledge and generalize across tasks~\cite{geng2025x}. Furthermore, architecture decoupling, such as using double image encoders, forces the cross-modal reasoning process to undergo additional decode-encode steps, which is inelegant and time-costing.

\begin{figure*}[t]
    \centering
    \includegraphics[width=0.98\linewidth]{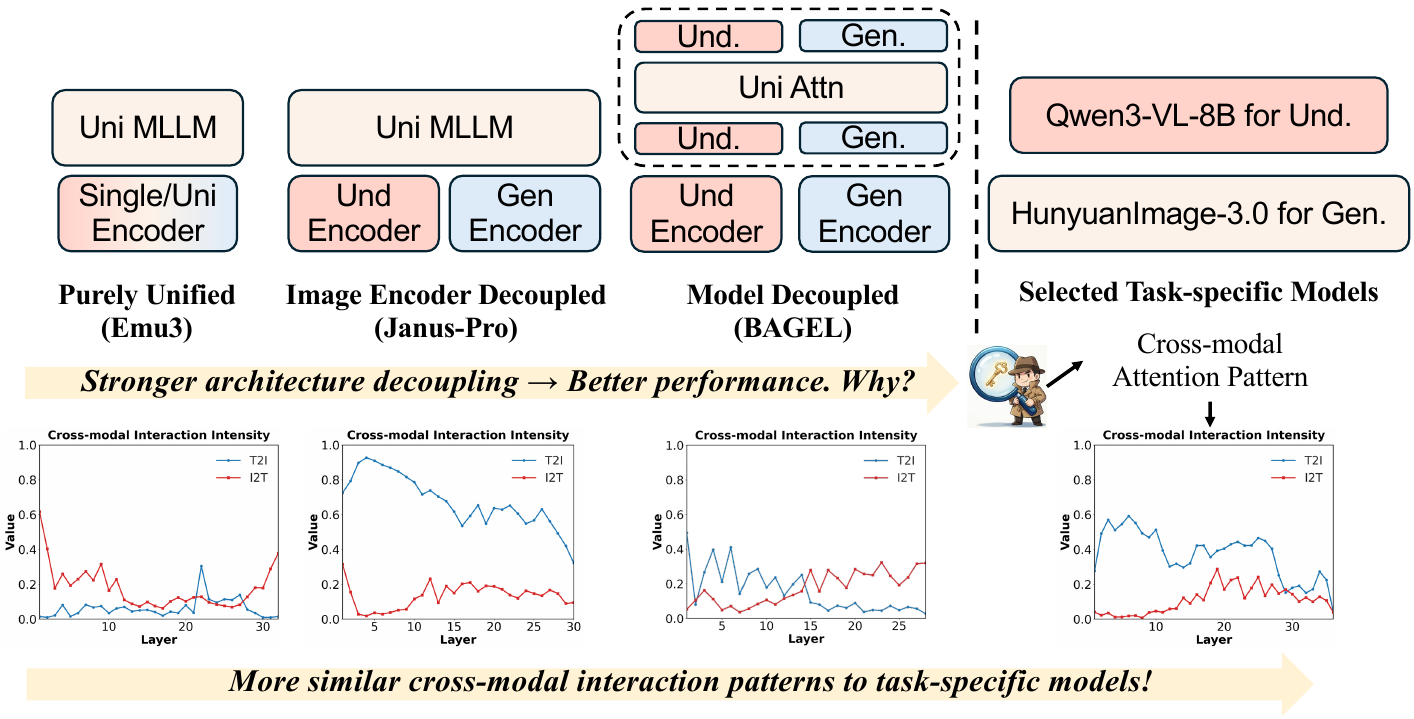}
    \vspace{-2mm}
    \caption{Various architectures of UMMs and its corresponding cross-modal interaction patterns. We arrange the models in order of increasing architecture decoupling. The row below illustrates the layer-wise cross-modal interaction intensity, with generation tasks shown in \textcolor{blue}{blue} and understanding tasks in \textcolor{red}{red}; higher values indicate stronger interaction. The last column corresponds to HunyuanImage-3.0 (\ie, \textbf{while HunyuanImage-3.0 is a purely unified model, we focus on its strong generation capabilities here.}) and Qwen3-VL-8B, representing the interaction behavior of current SOTA task-specific generation and understanding methods. We observe that as decoupling increases, the negative correlation in cross-modal interaction patterns between understanding and generation tasks persists, but these patterns increasingly resemble those of task-specific models, leading to improved performance.}
    \label{fig:moti}
    % \vspace{-2mm}
\end{figure*}

To maintain the original intent of UMM while narrowing its performance gap with decoupled models, we first conducted a detailed analysis of the underlying causes of this gap. Since the core of interaction between generation and understanding tasks lies in cross-modal information exchange, we focused our investigation on cross-modal interaction patterns (see \cref{sec:3.1} for details). As illustrated in \cref{fig:moti}, we first observe that regardless of architecture decoupling degrees, 
\textit{the two tasks show consistent negative correlation in cross-modal interaction patterns within each layer.} We further verified that this phenomenon is independent of input type or length; rather, the model dynamically allocates cross-modal representational weights within layers based on task requirements.
Moreover, as decoupling increases, the interaction patterns increasingly resemble those of task-specific models. This suggests that \textit{existing decoupled models do not eliminate the inherent conflict between tasks; instead, they make each task behave more like its single-task counterpart}, resulting in performance improvement.

Based on this observation, we propose Attention Interaction Alignment (AIA) loss, which will explicitly constrain the layer-wise cross-modal interaction intensity during training without architecture decoupling. Specifically, we first select models suitable as cross-modal interaction learning target. 
For the understanding target, we use Qwen3-VL-8B~\cite{bai2025qwen3vl}. 
For generation target, we refer to HunyuanImage-3.0~\cite{cao2025hunyuanimage}, which demonstrates superior generative capabilities despite its unified training paradigm.
For both understanding and generation tasks, we use 100 samples to extract the attention patterns at each layer and compute their average. Recognizing that attention patterns are closely linked to model architecture and pretraining, we apply the Huber loss to relax layer-wise attention constraints, enabling more flexible allocation of attention weights.

To validate the effectiveness of our proposed AIA loss, we adopt it to two degrees of decoupling methods (\eg, Emu3 for purely unified model, Janus-Pro for slight decoupling model). Experimental results demonstrate that our approach enhances both generation and understanding performance while narrowing the gap with more strongly decoupled methods. 
We further confirm that \textbf{the performance gain stems from the alignment of attention patterns rather than mere knowledge distillation, thereby corroborating our initial motivation.}
We highlight the main contributions of this paper below:
\begin{itemize}
    \item We provide the first mechanistic analysis of unified models with different decoupling architectures through cross-modal attention interaction intensity, revealing that decoupling does not resolve task conflicts but merely shifts attention patterns toward task-specific behaviors.
    \item We introduce AIA loss, a simple regularization to explicitly shift the cross-modal attention interaction patterns into the behavior of task-specific models without requiring architecture decoupling.
    \item Our method achieves significant improvements on widely-used generation and understanding benchmarks for both Emu3 and Janus-Pro.
\end{itemize}

\section{Related Works}
\label{sec:rw}
\subsection{Unified Multimodal Models (UMMs)}
In recent years, with the development of LLM and MLLM~\cite{Qwen2.5-VL, dubey2024llama, li2024llava, team2024qwen2, yang2025qwen3, liu2024deepseek,liu2025shotbench}, autoregressive architectures have been thoroughly explored, demonstrating exceptional capabilities in large models. Researchers are now considering whether MLLMs can be integrated with image generation to form a unified model, enabling automatic interleaved reasoning at the latent level. Initial unified MLLMs, such as Liquid~\cite{wu2024liquid}, Emu3~\cite{wang2024emu3}, and Chameleon~\cite{team2024chameleon}, adhered strictly to this path but used VAE~\cite{vae, vqgan, vqvae} as the image encoder, limiting their understanding capabilities. To address tokenizer representation conflicts, methods like TokenFlow~\cite{qu2025tokenflow}, Unitok~\cite{ma2025unitok}, Atoken~\cite{lu2025atoken}, UniLip~\cite{tang2025unilip} have constructed a unified tokenizer compatible with both understanding and generation (coarse-grained and fine-grained) feature requirements. VTP~\cite{vtp} further scales the unified tokenizer to a significant extent. The Janus series~\cite{januspro, wu2024janus, ma2024janusflow}, show-o~\cite{showo, xie2025showo2} series, and Transfusion~\cite{zhou2024transfusion, cao2025hunyuanimage} series have attempted to decouple the image encoder and training objectives for generation and understanding, achieving further performance improvements. However, due to conflicts between understanding and generation in the backbone network, performance remains constrained. To alleviate task conflicts, approaches like BAGEL~\cite{bagel} and OneCat~\cite{li2025onecat} have explored partially decoupled architectures (MOT, MOE), yielding promising results. Meanwhile, another group of researchers argues that generation hardly aids understanding, so Metaquery~\cite{metaquery}, OmniGen2~\cite{wu2025omnigen2}, UniWorld-V1~\cite{lin2025uniworld}, and Blip3-o~\cite{chen2025blip3} have chosen to fix the MLLM and solely optimize the diffusion head, achieving current state-of-the-art results. 

However, as model architectures become increasingly decoupled, starting from the image encoder, models can no longer achieve automatic interleaved reasoning in the latent space but must undergo an inelegant decode-encode process. Furthermore, as the backbone network becomes decoupled, the model's ability to handle understanding and generation in a unified manner is further weakened, let alone completely fixing the MLLM, which has deviated from the original intent of UMM. This paper aims to explore why architecture decoupling can alleviate conflicts and enhance performance, and then attempts to achieve similar effects within a fully unified architecture.

\subsection{Ultra Task-Specific Models}
Through extensive technological iterations, understanding tasks have developed into a nearly fixed model architecture. Models like Qwen~\cite{Qwen2.5-VL, team2024qwen2, yang2025qwen3, bai2025qwen3vl} and the LLaVA~\cite{dubey2024llama, li2024llava} series employ a semantic encoder combined with an autoregressive architecture, achieving outstanding results. In contrast, generation tasks lack a fixed model architecture. Early models such as SDXL~\cite{podell2023sdxl}, SD3~\cite{sd3}, and the FLUX~\cite{flux2024} series utilized CLIP~\cite{clip} as the text encoder within a pure diffusion~\cite{ddpm, ddim} framework, achieving high aesthetic quality but showing some limitations in instruction compliance. SimpleAR~\cite{wang2025simplear} explored image generation using LLM as the base model within a purely autoregressive architecture, but the inherent information loss in discrete representations resulted in images with a pronounced blur. Qwen-Image~\cite{qwenimage}, Longcat-Image~\cite{longcatimage} and Z-Image~\cite{zimage} combined the strengths of both approaches by replacing CLIP with MLLM and integrating a diffusion head after the MLLM, achieving excellent results in both instruction compliance and aesthetic quality. HunyuanImage-3.0~\cite{cao2025hunyuanimage} is trained in a unified manner while showing great generation ability. In this paper, we select the most performant models in understanding (Qwen3-VL) and generation (HunyuanImage-3.0) tasks as references for cross-modal interaction patterns, thereby enhancing the reliability of the conclusions.

\subsection{Knowledge Distillation}
Knowledge distillation~\cite{hinton2015distilling}, initially proposed by Hinton, is a technique that transfers knowledge from a teacher model to a student model. FitNets~\cite{fitnet} further advances this approach by incorporating intermediate features to better align teacher and student models. \textbf{Our AIA loss differs fundamentally from traditional output- or feature-level knowledge distillation} in two key aspects. 1) We rely solely on the layer-wise cross-modal interaction intensity. This is a pre-computed statistic that involves no specific feature and eliminates the need for a concurrent teacher model during training. 2) Standard distillation inherits the severe conflict between generation and understanding inherent in UMMs. Consequently, imposing rigid constraints proves counterproductive and degrades performance, as we verify in \Cref{tab:ablation}. AIA emerges as an intuitive and effective solution derived directly from our empirical insights into these conflicts.

\begin{wraptable}{r}{0.55\textwidth}
\vspace{-4mm}
\centering
\caption{Standard deviation across 100 samples for each model.}
\resizebox{0.5\textwidth}{!}{
\begin{tabular}{c@{\hspace{1em}}c@{\hspace{1em}}c@{\hspace{1em}}cccc}
\toprule
\textbf{Method} &
\textbf{Emu3} & \textbf{Janus-Pro} & \textbf{BAGEL} & \textbf{Task-Specific} \\
\midrule[0.12em]
Std & 0.13 & 0.02 & 0.03 & 0.1 \\
\bottomrule
\end{tabular}}
\label{tab:std}
\vspace{-2mm}
\end{wraptable}

\section{Attention Interaction Alignment}
% \vspace{-2mm}
In this section, we first analyze that why model decoupling will alleviate the task conflicts and improve the performance based on cross-modal attention interaction pattern. We observe that regardless of architecture decoupling degrees, different tasks induce mutually exclusive cross-modal attention patterns across various layers, but push the pattern into task-specific model types. Based on this, we propose Attention Interaction Alignment loss to constrain the attention pattern based on the task-specific one during training.

\subsection{Analysis of Cross-Modal Attention Interaction Pattern in Various Model Architectures}
\label{sec:3.1}
\vspace{-2mm}
\mbox{}
\begin{wrapfigure}{r}{0.5\textwidth}
\vspace{-7mm}
  \centering
  \includegraphics[width=0.48\textwidth]{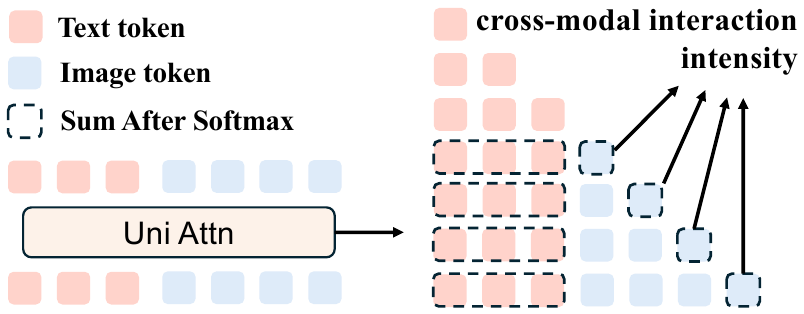}
  % \vspace{-2mm}
  \caption{The pipeline of cross-modal interaction intensity calculation. We take text-to-image as an example, for each row, we compute the sum of all text token values, then average across all image tokens to obtain the intensity.}
  \label{fig:arch}
\vspace{-10mm}
\end{wrapfigure}
\noindent\textbf{Method.} We take text-to-image generation as an 
example. As shown in \cref{fig:arch}, 
after obtaining the attention map from the softmax operation, we calculate the sum of all text tokens in each row of the attention map. Then, we compute the 
average over all layers, all image tokens and all attention heads. This value represents the cross-modal interaction intensity between images and text for the given sample. The formula is as follows:
\begin{equation}
I_l = \frac{1}{H \times Q} \sum_{h=1}^{H} \sum_{q=1}^{Q} \sum_{k=1}^{K} \text{Attn}_{l}(h, q, k),
\end{equation}
where I means the intensity, l is the layer index, H, Q, K represent head, query, key numbers respectively. The average results of 100 samples are shown in \cref{fig:moti}. In \Cref{tab:std}, we further validate that the attention pattern of each model is independent of input length and type (\ie, consisting of prompts like text rendering, short and dense caption and questions like caption, choices, \etc) by computing the standard deviation for each model.

\noindent\textbf{Rationality of task-specific attention pattern.} \cref{fig:moti} shows that Qwen3-VL~\cite{bai2025qwen3vl} consistently exhibits low attention to image tokens, which aligns with the motivation behind token pruning methods~\cite{wang2025sparsemm, zhang2024sparsevlm} in current understanding tasks. For generation, HunyuanImage-3.0~\cite{cao2025hunyuanimage} maintains around 40\% attention to text tokens in the first 80 layers, with a sharp decline in the final layers. This pattern matches the common consensus in generative models~\cite{repa}: shallow layers focus on building semantic representations and thus attend more to high-level text features, while the final layers shift toward pixel-level image features, resulting in reduced attention to text. The decrease in text attention in deeper layers further supports the claim in RecA~\cite{reca} that textual information is insufficient at these stages. Except for Emu3~\cite{wang2024emu3}, all other architectures follow the single-task trend, indicating that unified models tend to learn in a task-specific manner once their architectures are decoupled.

\noindent\textbf{Why model decoupling improves performance.} From \Cref{tab:std}, we can first rule out the influence of input properties on cross-modal attention patterns. \cref{fig:moti} shows that the interaction curves for the two tasks have negative correlation, indicating that, after unified training, the inherent conflict between tasks forces the model to allocate attention weights to different tasks at different layers. This mechanism helps the model self-mitigate cross-task interference, and the resulting allocation pattern becomes fixed. Furthermore, \cref{fig:moti} demonstrates that this negative correlation persists regardless of the degree of model decoupling. As the model becomes more decoupled, the attention interaction patterns increasingly resemble those of task-specific models, leading to improved performance. Specifically, when performing understanding tasks, Janus-Pro exhibits relatively high attention to image tokens in the initial layers, which limits its understanding capability. For Emu3, both generation and understanding tasks deviate significantly from the typical single-task interaction patterns, highlighting the inherent difficulty of fully unified learning within an autoregressive architecture.

\subsection{Attention Interaction Alignment Loss}
Based on the observation above, we propose attention interaction alignment loss, which will constrain the attention patterns explicitly during training. Specifically, we use the layer-wise intensity from the task-specific models in \cref{fig:moti} as learning targets, termed $T_l$. However, since the attention curves are fixed throughout network training, it is unsuitable to apply overly strict constraints for supervision. Therefore, we divide the values into several sub-stages according to their magnitudes and further relax the absolute constraint on individual values using the Huber loss. Taking Emu3~\cite{wang2024emu3} as an example, the specific formula is as follows: 

\begin{equation}
\mathcal{L}_{\text{AIA}} = \frac{1}{L} \sum_{l=1}^{L} 
\begin{cases}
\frac{1}{2}(I_l - T_l)^2, & \text{if } |I_l - T_l| \leq \delta_l \\
\delta_l \cdot |I_l - T_l| - \frac{1}{2} \delta_l^2, & \text{otherwise}
\end{cases}
\end{equation}

where $I_l$ denotes the cross-modal intensity at layer $l$, and for the layer-wise 
\begin{wraptable}{r}{0.57\textwidth}
\vspace{-2mm}
\centering
\caption{The hyper-parameter of Huber loss in Emu3 and Janus-Pro.}
\resizebox{0.5\textwidth}{!}{
\begin{tabular}{ccc}
\toprule[0.12em]
\textbf{Layer Range} & \textbf{Generation} $(\delta_l, T_l)$ & \textbf{Understanding} $(\delta_l, T_l)$ \\
\midrule[0.12em]
$0 \leq l < 10$ & $(0.2, 0.4)$ & $(0.05, 0.1)$ \\
$10 \leq l < 20$ & $(0.1, 0.4)$ & $(0.05, 0.15)$ \\
$20 \leq l < 25$ & $(0.1, 0.4)$ & $(0.05, 0.3)$ \\
$25 \leq l \leq L$ & $(0.05, 0.2)$ & $(0.05, 0.3)$ \\
$l > L$ & $(0.05, 0.2)$ & $(0.05, 0.2)$ \\
\bottomrule[0.1em]
\end{tabular}}
\label{tab:param}
\vspace{-2mm}
\end{wraptable}
target boundary $T_l$ and Huber threshold $\delta_l$, we show the configuration in \Cref{tab:param} (L is set to 30 in Emu3 and 29 in Janus-Pro). We assign distinct values of T to different layers of the target model while dynamically adjusting $\delta$ based on the layer-wise similarity between Emu3, Janus-Pro and target models. This adaptive strategy prevents performance degradation caused by excessively rigid constraints. The AIA loss will combine with next-token-prediction loss, term $\mathcal{L}_{\text{NTP}}$ as:
\begin{equation}
    \mathcal{L} = \mathcal{L}_{\text{NTP}} + \lambda \times \mathcal{L}_{\text{AIA}},
\end{equation}
where $\lambda$ is set to 40 in the experiment by default.

\subsection{Training Details Under Different Architectures}
\label{sec:3.3}
We incorporate the AIA loss into Emu3 during the supervised fine-tuning (SFT) stage and into Janus-Pro during the post-training stage to demonstrate the effectiveness of our method across different scenarios and explore the challenges of integrating our loss at various training phases.

\begin{figure*}[t]
    \centering
    \includegraphics[width=0.98\linewidth]{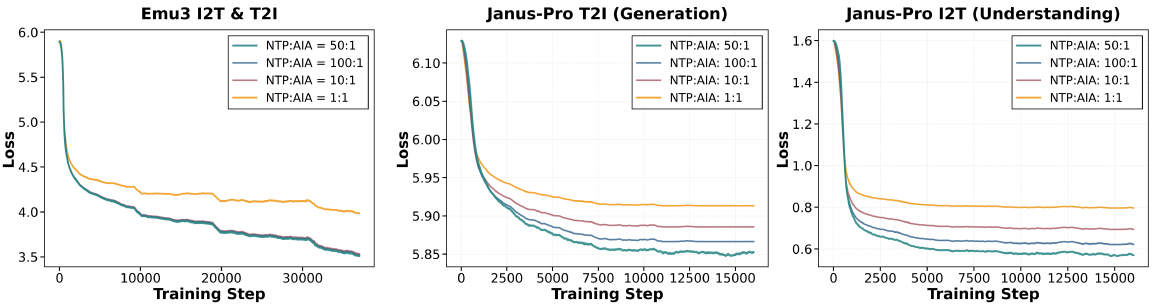}
    % \vspace{-2mm}
    \caption{Training loss curve of Emu3 and Janus-Pro under various AIA coefficient. NTP and AIA means next-token-prediction and attention interaction alignment loss respectively. Note that we only show the NTP loss curve (excluding AIA loss) as it serves as the primary indicator of final performance and the periodic drops in the Emu3 loss curve are due to learning rate schedule.
}
    \label{fig:loss}
    % \vspace{-2mm}
\end{figure*}

\noindent\textbf{Emu3.} We load the pretrained (PT) weights of Emu3 and perform SFT training using our own data. Note that Emu3 only employs unified training at this stage; the results reported in the original paper were obtained through separate training. Since Emu3's performance at the pretraining stage is relatively poor, as shown in \cref{fig:loss}, we incorporate the AIA loss with varying weights during the SFT stage and observe that the NTP loss convergence trends remain nearly identical. This indicates that, when the learning target is set appropriately, incorporating the AIA loss during SFT does not significantly affect the model's pretrained knowledge.

\noindent\textbf{Janus-Pro.} Since Janus-Pro only provides the final SFT weights, we perform post-training on this basis. When data quality does not differ significantly, this greatly increases the difficulty of tuning, as the model's distribution is already highly fixed at this stage (\ie, a point further supported by the variance comparison between Emu3 and Janus-Pro in \Cref{tab:std}). However, this also aligns more closely with realistic settings: \textit{how to perform fine-grained tuning when only the final weights are accessible and attention distribution adjustment is required}. As shown in \cref{fig:loss}, the model is highly sensitive to the weight of the AIA loss at this stage, yet it can still achieve the desired effect when $\lambda$ is set appropriately. We further validate this in \cref{sec:ablat}.

\begin{table*}[t]
    \centering
    \caption{System-level comparison on widely used image understanding and generation benchmarks. $\dagger$ means the result is re-implemented by ourselves. Types represent whether the model uses diffusion, autoregressive, or masked prediction for training. \textcolor{gray}{(Gray)} means the result reported in original papers while nobody can re-implemented.}
    \vspace{-2mm}
    \resizebox{1\columnwidth}{!}{
        \begin{tabular}{llcccccccccc}
        \toprule[0.15em]
        \multicolumn{2}{l}{\multirow{2}{*}{\textbf{Method}}} & \multirow{2}{*}{\textbf{Params}} & \multirow{2}{*}{\textbf{Types}} & \multicolumn{6}{c}{\textbf{Image Understanding}} & 
        \multicolumn{2}{c}{\textbf{Image Generation}} \\
        \multicolumn{2}{c} {} & {} & {} & MMMU & MMBench & MMVP & MMVet & POPE & MME-P & GenEval & DPG  \\
        \midrule[0.1em]
        \rowcolor{gray!10}
        \multicolumn{12}{c}{\textbf{\textit{Gen. Only}}} \\
        \midrule[0.1em]
        \multicolumn{2}{l}{SDXL~\cite{podell2023sdxl}} & - & Diff & & & - & & & & 0.55 & 74.65 \\
        \multicolumn{2}{l}{SD3-medium~\cite{sd3}} & 2B & Diff & & & - & & & & 0.74 & 84.08 \\
        \multicolumn{2}{l}{Infinity~\cite{han2025infinity}} & 2B & VAR & & & - & & & & 0.73 & 83.50 \\
        \multicolumn{2}{l}{Infinity~\cite{han2025infinity}} & 8B & VAR & & & - & & & & 0.79 & 86.60 \\
        \multicolumn{2}{l}{FLUX.1-dev~\cite{flux2024}} & 12B & Diff & & & - & & & & 0.82 & 84.00 \\
        \multicolumn{2}{l}{Emu3-Gen~\cite{wang2024emu3}} & 8B & AR & & & - & & & & 0.66 & 80.60 \\
        \multicolumn{2}{l}{Qwen-Image~\cite{qwenimage}} & 7B+20B & AR+Diff & & & - & & & & 0.87 & 88.32 \\
        \midrule[0.1em]
        \rowcolor{gray!10}
        \multicolumn{12}{c}{\textbf{\textit{Und. Only}}} \\
        \midrule[0.1em]
        \multicolumn{2}{l}{Emu3-Chat~\cite{wang2024emu3}} & 8B & AR & 31.6 & 58.5 & 36.6 & 37.2 & 85.2 & 1244 & - & - \\
        \multicolumn{2}{l}{Qwen2.5-VL~\cite{Qwen2.5-VL}} & 3B & AR & 53.1 & 79.1 & - & 61.8 & - & - & - & - \\
        \multicolumn{2}{l}{Qwen2.5-VL~\cite{Qwen2.5-VL}} & 7B & AR & 58.6 & 83.5 & - & 66.6 & - & 1685 & - & - \\
        \multicolumn{2}{l}{InternVL2.5~\cite{chen2024internvl}} & 8B & AR & 56.2 & 84.6 & - & 62.8 & 90.6 & - & - & - \\
        \multicolumn{2}{l}{InternVL3~\cite{zhu2025internvl3}} & 8B & AR & 62.7 & 83.4 & - & 81.3 & 91.1 & - & - & - \\
        \multicolumn{2}{l}{Qwen3-VL~\cite{bai2025qwen3vl}} & 8B & AR & 69.6 & 85.0 & - & - & - & - & - & - \\
        \midrule[0.1em]
        \rowcolor{gray!10}
        \multicolumn{12}{c}{\textbf{\textit{Uni. Frozen MLLM}}} \\
        \midrule[0.1em]
        \multicolumn{2}{l}{MetaQuery-XL~\cite{metaquery}} & 7B+1.6B & AR+Diff & 58.6 & 83.5 & - & 66.6 & - & 1685 & 0.80 & 82.05 \\
        \multicolumn{2}{l}{Blip3-o~\cite{chen2025blip3}} & 7B+1.4B & AR+Diff & 58.6 & 83.5 & - & 66.6 & - & 1685 & 0.84 & 81.60 \\
        \multicolumn{2}{l}{UniWorld-V1~\cite{lin2025uniworld}} & 7B+12B & AR+Diff & 58.6 & 83.5 & - & 67.1 & - & 1685 & 0.84 & 81.38 \\
        \multicolumn{2}{l}{OmniGen2~\cite{wu2025omnigen2}} & 3B+4B & AR+Diff & 53.1 & 79.1 & - & 61.8 & - & - & 0.86 & 83.57 \\
        \midrule[0.1em]
        \rowcolor{gray!10}
        \multicolumn{12}{c}{\textbf{\textit{Uni. MoE/MoT Arch}}} \\
        \midrule[0.1em]
        \multicolumn{2}{l}{BAGEL~\cite{bagel}} & 7B+7B & AR+Diff & 55.3 & 85.0 & 69.3 & 67.2 & - & 1687 & 0.88 & 85.07 \\
        \multicolumn{2}{l}{OneCat~\cite{li2025onecat}} & 3B+3B+3B & AR+VAR & 41.9 & 78.8 & - & 52.2 & - & 1630 & 0.90 & 84.53 \\
        \midrule[0.1em]
        \rowcolor{gray!10}
        \multicolumn{12}{c}{\textbf{\textit{Uni. Double Image Encoders / Training Objectives}}} \\
        \midrule[0.1em]
        \multicolumn{2}{l}{Show-o~\cite{showo}} & 1.3B & AR+Mask & 26.7 & - & - & - & 80.0 & 1097 & 0.69 & 67.27 \\
        \multicolumn{2}{l}{Show-o2~\cite{xie2025showo2}} & 7B & AR+Diff & \textbf{48.9} & \textbf{79.3} & - & - & - & \underline{1621} & 0.76 & \textbf{86.14} \\
        \multicolumn{2}{l}{Janus-Pro~\cite{januspro}} & 7B & AR & 41.0 & 65.54 \textcolor{gray}{(79.2)} & 47.3 & \textbf{50.0} & 87.4 & 1567 & \underline{0.80} & 84.19 \\
        \multicolumn{2}{l}{\textbf{Janus-Pro + AIA (Ours)}} & 7B & AR & \underline{42.1} & \underline{75.6} & \textbf{48.0} & \underline{49.8} & \textbf{89.8} & \textbf{1656} & \textbf{0.81} & \underline{84.49} \\
        \midrule[0.1em]
        \rowcolor{gray!10}
        \multicolumn{12}{c}{\textbf{\textit{Uni. Purely}}} \\
        \midrule[0.1em]
        \multicolumn{2}{l}{Chameleon~\cite{team2024chameleon}} & 7B & AR & 28.4 & 35.7 & - & 8.3 & - & - & 0.39 & - \\
        \multicolumn{2}{l}{VILA-U~\cite{wu2024vila}} & 7B & AR & \underline{32.2} & \textbf{66.6} & \textbf{22.0} & \textbf{27.7} & \textbf{83.9} & \textbf{1336} & 0.39 & 72.48 \\
        \multicolumn{2}{l}{Emu3~\cite{wang2024emu3} $\dagger$} & 8B & AR & 31.6 & 61.4 & 8.7 & 15.1 & 77.3 & 910 & \underline{0.60} & \underline{79.24} \\
        \multicolumn{2}{l}{\textbf{Emu3 + AIA (Ours)}} & 8B & AR & \textbf{35.7} & \underline{64.8} & \underline{10.8} & \underline{18.7} & \underline{82.7} & \underline{1084} & \textbf{0.67} & \textbf{81.20} \\
        \bottomrule[0.1em]
    \end{tabular}{}
    }
    \vspace{-2mm}
    \label{tab:main}
\end{table*}

\begin{table*}[t]
    \centering
    \caption{Quantitative comparison for the ablation study about the data quality, AIA loss, atention pattern, $\lambda$, and data sampling ratio.}
    \vspace{-2mm}
    \resizebox{1\columnwidth}{!}{
        \begin{tabular}{llcccccccc}
        \toprule[0.15em]
        \multicolumn{2}{l}{\multirow{2}{*}{\textbf{Method}}} & \multicolumn{6}{c}{\textbf{Image Understanding}} & 
        \multicolumn{2}{c}{\textbf{Image Generation}} \\
        \multicolumn{2}{c} {} & MMMU & MMBench & MMVP & MMVet & POPE & MME-P & GenEval & DPG  \\
        \midrule[0.1em]
        \multicolumn{4}{l}{\textit{Data Quality and The effectiveness of AIA Loss}} \\
        \midrule[0.1em]
        \multicolumn{2}{l}{\textbf{Emu3 + AIA (Final)}} & \textbf{35.7} & \textbf{64.8} & \textbf{10.8} & \textbf{18.7} & \textbf{82.7} & \textbf{1084} & \textbf{0.67} & \textbf{81.20} \\
        \multicolumn{2}{l}{w/o AIA (baseline)} & 31.6 & 61.4 & 8.7 & 15.1 & 77.3 & 910 & 0.60 & 79.24 \\
        \multicolumn{2}{l}{\textbf{Janus-Pro + AIA (Final})} & \textbf{42.1} & \textbf{75.6} & \textbf{48.0} & \textbf{49.8} & \textbf{89.8} & \textbf{1656.4} & \textbf{0.81} & \textbf{84.49} \\
        \multicolumn{2}{l}{w/o stage-level intensity} & 40.2 & 67.4 & 43.1 & 44.0 & 87.6 & 1543.2 & 0.79 & 83.79 \\
        \multicolumn{2}{l}{w/o Huber} & 41.2 & 73.1 & 47.6 & 47.3 & 88.2 & 1613.9 & 0.80 & 84.25 \\
        \multicolumn{2}{l}{w/o AIA (baseline)} & 40.7 & 71.5 & 47.3 & 49.2 & 88.1 & 1593.1 & 0.80 & 84.19 \\
        \midrule[0.1em]
        \multicolumn{4}{l}{\textit{AIA vs. Knowledge Distillation (Janus-Pro)}} \\
        \midrule[0.1em]
        \multicolumn{2}{l}{\textbf{AIA (ours)}} & \textbf{42.1} & \textbf{75.6} & \textbf{48.0} & \textbf{49.8} & \textbf{89.8} & \textbf{1656.4} & \textbf{0.81} & \textbf{84.49}\\
        \multicolumn{2}{l}{Feature-level} & 32.5 & 58.6 & 41.0 & 40.3 & 85.5 & 1485.4 & 0.72 & 81.97 \\
        \multicolumn{2}{l}{Output-level} & 40.2 & 68.4 & 47.3 & 46.9 & 87.7 & 1603.7 & 0.79 & 83.97 \\
        \midrule[0.1em]
        \multicolumn{4}{l}{\textit{Attention Pattern Selection (Janus-Pro)}} \\
        \midrule[0.1em]
        \multicolumn{2}{l}{Qwen3-VL+FLUX} & 40.2 & 72.1 & 45.5 & 46.0 & 87.5 & 1555.1 & 0.76 & 83.56 \\
        \multicolumn{2}{l}{Qwen3-VL+SimpleAR} 
        & 40.3 & 72.5 & 44.1 & 46.6 & 88.0 & 1546.7 & 0.75 & 82.89 \\
        \multicolumn{2}{l}{Qwen3-VL+HunyuanImage-3.0} & \textbf{42.1} & \textbf{75.6} & \textbf{48.0} & \textbf{49.8} & \textbf{89.8} & \textbf{1656.4} & \textbf{0.81} & \textbf{84.49}\\
        \multicolumn{2}{l}{Qwen3-VL+Qwen-Image} & 41.5 & 74.3 & 47.3 & 48.6 & 89.5 & 1623.4 & 0.80 & 84.12 \\
        \midrule[0.1em]
        \multicolumn{4}{l}{\textit{$\lambda$ Selection (Janus-Pro)}} \\
        \midrule[0.1em]
        \multicolumn{2}{l}{NTP:AIA: 1:1} & 37.2 & 62.9 & 44.5 & 44.0 & 87.3 & 1498.9 & 0.77 & 83.35\\
        \multicolumn{2}{l}{NTP:AIA: 10:1} & 40.5 & 71.1 & 46.6 & 48.2 & 87.2 & 1545.6 & 0.80 & 83.26 \\
        \multicolumn{2}{l}{NTP:AIA: 50:1} & \textbf{42.1} & \textbf{75.6} & \textbf{48.0} & \textbf{49.8} & \textbf{89.8} & \textbf{1656.4} & \textbf{0.81} & \textbf{84.49} \\
        \multicolumn{2}{l}{NTP:AIA: 100:1} & 41.3 & 71.3 & 47.3 & 49.7 & 87.9 & 1595.2 & 0.80 & 84.11\\
        \midrule[0.1em]
        \multicolumn{4}{l}{\textit{Data Sampling Ratio (Janus-Pro)}} \\
        \midrule[0.1em]
        \multicolumn{2}{l}{Gen:Und: 1:1} & \textbf{42.1} & \textbf{75.6} & \textbf{48.0} & \textbf{49.8} & \textbf{89.8} & \textbf{1656.4} & \textbf{0.81} & \textbf{84.49} \\
        \multicolumn{2}{l}{Gen:Und: 2:1} & 41.8 & 74.2 & 46.6 & 49.3 & 89.1 & 1643.6 & 0.81 & 84.23\\
        \multicolumn{2}{l}{Gen:Und: 4:1} & 41.2 & 72.8 & 47.3 & 48.5 & 88.3 & 1621.4 & 0.80 & 83.87\\
        \multicolumn{2}{l}{Gen:Und: 1:2} & 42.0 & 74.8 & 47.3 & 49.3 & 89.3 & 1645.9 & 0.79 & 83.12\\
        \bottomrule[0.1em]
    \end{tabular}{}
    }
    \vspace{-2mm}
    \label{tab:ablation}
\end{table*}

\section{Experiment}
\subsection{Experimental Setup}
\noindent\textbf{Datasets.} We primarily use open-source image generation and understanding datasets for training, including ShareGPT-4V~\cite{chen2025sharegpt}, BLIP3-o~\cite{chen2025blip3}, and OpenSora~\cite{zheng2024open} for generation, and LLaVA-OneVision-1.0~\cite{li2024llava}, Mammoth-VL~\cite{guo2024mammoth} for understanding, resulting in 1.5M samples for each task. Since the quality of these datasets is not as high as that of Janus-Pro's data, we incorporate 200K internal samples to align the data quality.

\noindent\textbf{Implementation Details.} For Emu3, we proportionally resize images to approximately 720×720 resolution for both understanding and generation tasks. The understanding and generation data are balanced at a 1:1 ratio. Under the DeepSpeed ZeRO-3 framework, the entire training process for the 8B model took approximately 10 days on a cluster of 8 nodes, each equipped with 8 NVIDIA H800 (80GB) GPUs. For Janus-Pro, we proportionally resize images to approximately 384×384 resolution for both understanding and generation tasks. The understanding and generation data are balanced at a 1:1 ratio. Under the FSDP framework, the entire training process for the 7B model took approximately 1 day on a cluster of 8 nodes, each equipped with 8 NVIDIA H800 (80GB) GPUs.

\subsection{Evaluation on the benchmarks}
\noindent\textbf{Multimodal Understanding.} We evaluate our model on six widely recognized benchmarks—MME~\cite{fu2023mme}, MMBench (1.0-EN)~\cite{liu2024mmbench}, MMVet~\cite{mmvet}, MMMU~\cite{yue2024mmmu}, POPE~\cite{pope}, and MMVP~\cite{mmvp}—which together form a compact yet thorough assessment framework encompassing perception, cognition, and multimodal reasoning, with robust capability to distinguish performance differences among leading models. The result is shown in \Cref{tab:main}, 
within each decoupling degree, we achieve state-of-the-art results under the same training configuration, and narrow the performance gap with models employing higher decoupling degrees or different training paradigms.

\noindent\textbf{Text-to-Image Generation.} We follow Janus-Pro~\cite{januspro} and report results on widely used image generation benchmarks Geneval~\cite{ghosh2023geneval} and DPG-Bench~\cite{dpg}. As shown in \Cref{tab:main}, equipping with the AIA loss regulation, we improve the performance of Janus-Pro and Emu3 and narrow the gap with models with higher decoupling degrees.

\begin{wraptable}{r}{0.5\textwidth}
\vspace{-4mm}
\centering
\caption{Quantitative comparison on Real-Unify.}
\resizebox{0.46\textwidth}{!}{
\begin{tabular}{c@{\hspace{1em}}c@{\hspace{1em}}c@{\hspace{1em}}cccc}
\toprule
\textbf{Method} & \textbf{Janus-Pro} & \textbf{Janus-Pro+AIA} \\
\midrule[0.12em]
Step-wise results & 25.2 & 27.8 \\
\bottomrule
\end{tabular}}
\label{tab:interleave}
\vspace{-2mm}
\end{wraptable}
\noindent\textbf{Interleave Reasoning.} Here we show the improvement on interleave reasoning ability. Since current fully unified models and encoder-decoupled methods are still in the early stages of development and have not yet demonstrated mature 
interleaved reasoning abilities (\ie, evidenced by the fact that no experiments regarding these models are reported in the interleaved reasoning benchmark Uni-MMMU~\cite{unimmmu}). We primarily validate our method on the ``Understanding aids Generation'' metric (which serves as a step toward interleaved reasoning) from RealUnify~\cite{shi2025realunify} on Janus-Pro. The result in \Cref{tab:interleave} shows that AIA improves the interleave reasoning performance compared with Janus-Pro.

\subsection{Ablation Study}
\label{sec:ablat}
Due to the slow training speed of Emu3, we primarily conduct ablation studies on Janus-Pro. As mentioned in \cref{sec:3.3}, this represents a more challenging setting, thereby better demonstrating the effectiveness of our method.

\noindent\textbf{Data Quality Analysis.} Since the training data of Janus-Pro is not publicly available, we first assess the quality of our own dataset to validate our method's effectiveness. As shown in \Cref{tab:ablation} (\ie, w/o AIA), further fine-tuning Janus-Pro with our data yields comparable performance to the original model in both understanding and generation, indicating that \textit{our data quality itself does not introduce performance gains}. Note that the Emu3 results (w/o AIA) cannot be directly compared with those reported in the original Emu3 paper, as their performance was obtained through single-task SFT. The relatively low unified performance, particularly in understanding, highlights the inherent limitations of using a VAE as the image encoder and the challenges of purely unified training.

\begin{figure*}[t]
    \centering
    \includegraphics[width=0.98\linewidth]{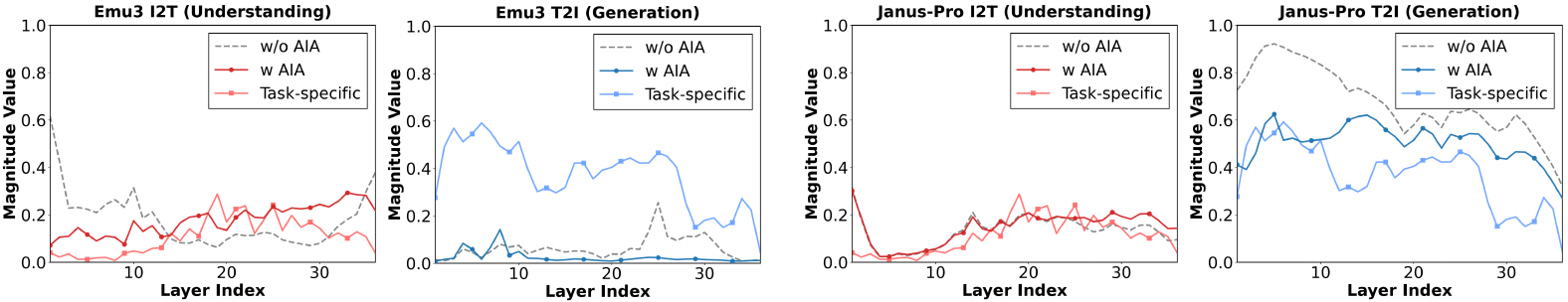}
    \vspace{-2mm}
    \caption{Visualization of cross-modal attention patterns modification after AIA training. Task-specific models are Qwen3-VL-8B for understanding and HunyuanImage-3.0 for generation, with understanding tasks shown in \textcolor{red}{red} and generation tasks in \textcolor{blue}{blue}.}
    \label{fig:aialoss}
    \vspace{-4mm}
\end{figure*}

\noindent\textbf{The Effectiveness of AIA loss.} The result in \Cref{tab:ablation} shows that incorporating our AIA loss leads to improved performance in both understanding and generation, with earlier integration (\eg, during the Emu3 SFT stage) yielding greater improvements. \cref{fig:aialoss} further illustrates the changes in cross-modal attention patterns after applying the AIA loss. With AIA regularization, the attention patterns of both Emu3 and Janus-Pro shift closer to those of task-specific models, confirming that aligning attention patterns toward task-specific behaviors indeed enhances model performance, and that the AIA loss effectively improves results while reshaping attention patterns. However, it could also be observed that when adding AIA loss to purly unified model architecture Emu3, the cross-modal attention pattern is more harder to change (\ie, the generation attention pattern in Emu3 merely captures the correct directional trend across layers, but the actual values are still incorrect).

Then we validate the effectiveness of huber loss and stage-level intensity. Both components are designed to relax overly strict attention constraints. As demonstrated in \Cref{tab:ablation}, eliminating either component results in performance degradation below the baseline, suggesting that overly rigid attention constraints impede the training process. The Huber loss and stage-level intensity provide coarse optimization targets while allowing the model flexibility for self-adjustment, effectively addressing this issue and validating the effectiveness of AIA loss.

\noindent\textbf{Performance Gain From Knowledge Distillation or AIA?} Although we rely on coarse-grained cross-modal attention statistics from task-specific models (see \cref{sec:3.1} for calculation details), a fundamental question arises regarding the source of our gains. We need to determine whether the performance improvement stems from general knowledge distillation or specifically from the alignment of 
attention patterns. This distinction is crucial for verifying that attention interaction patterns are indeed the primary cause of performance disparities across different architectures. To investigate this, \Cref{tab:ablation} compares our AIA method against standard feature-based and output-based distillation on Janus-Pro. 

The results indicate that \textbf{naively applying traditional knowledge distillation fails to improve performance in UMM training.} This is due to the inherent conflict between generation and understanding tasks, which persists even during distillation. In contrast, constraining attention interaction patterns is the only effective strategy for performance gains. This finding validates our motivation that architecture decoupling improves performance by shifting attention patterns.

\begin{figure*}[t]
    \centering
    \includegraphics[width=0.98\linewidth]{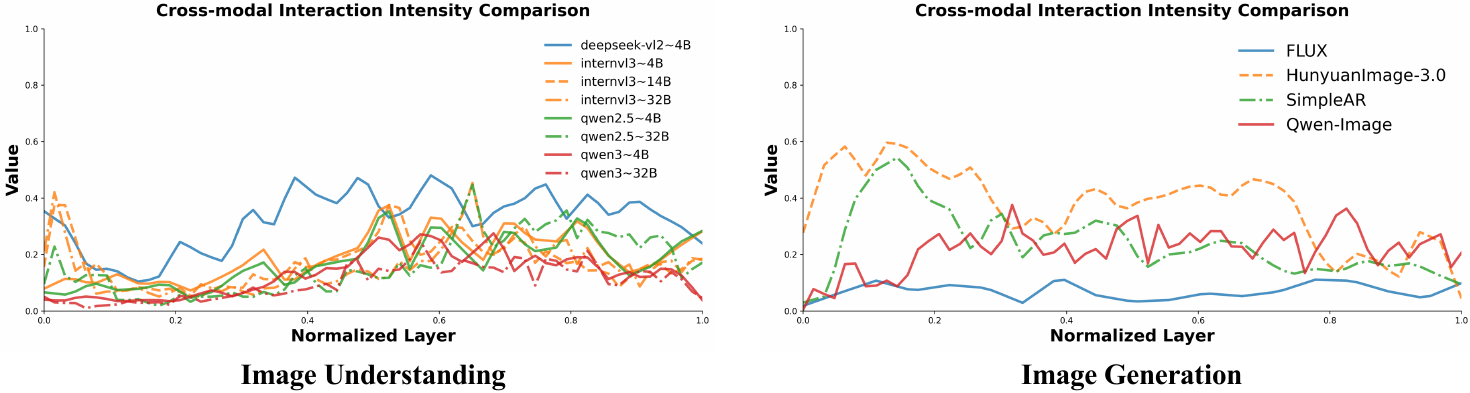}
    % \vspace{-2mm}
    \caption{Cross-Modal Attention Patterns Visualization of Different Single-Task Models.}
    \label{fig:cross-attn}
    \vspace{-4mm}
\end{figure*}

\noindent\textbf{Task-specific Attention Patterns Selection.} To identify the most suitable task-specific attention patterns as training targets for unified models, we compare Deepseek-VL2~\cite{liu2024deepseek}, the InternVL~\cite{chen2024internvl, zhu2025internvl3} series, and the Qwen~\cite{Qwen2.5-VL,bai2025qwen3vl} series for understanding tasks, and FLUX~\cite{flux2024}, SimpleAR~\cite{wang2025simplear}, Qwen-Image~\cite{qwenimage} and HunyuanImage-3.0~\cite{cao2025hunyuanimage} for generation tasks. \cref{fig:cross-attn} illustrates the corresponding attention patterns of these models. We observe that attention patterns for understanding tasks remain highly consistent across different models, whereas those for generation tasks exhibit significant variations. Consequently, we select Qwen3-VL-8B as the target for understanding and combine it with attention patterns from four generation models. \Cref{tab:ablation} shows that the attention pattern from HunyuanImage-3.0 achieves the best performance as a learning target, 
while that from SimpleAR and FLUX yield mediocre results. This suggests that the quality of the generated attention patterns depends on the model's inherent capacity, rather than the choice of attention mechanism (causal vs. bidirectional).
We also argue that HunyuanImage-3.0 may not represent the optimal choice, as its training undergoes reinforcement learning, making its attention patterns potentially suboptimal for the PT or SFT stages of unified model training.

\noindent\textbf{$\lambda$ Selection.} Here, we analyze the coefficient of the AIA loss during fine-tuning on Janus-Pro. We show the results in \Cref{tab:ablation}, where NTP:AIA denotes the loss weight ratio between the two objectives. By modifying attention patterns in a model with well-established pretrained knowledge is highly sensitive to the loss coefficient (\ie, also validated in \cref{fig:loss}). The coefficient must be carefully balanced—strong enough to influence model optimization, yet not so dominant as to disrupt the existing knowledge.

\noindent\textbf{Is Data Sampling Ratios still Matter with AIA loss?} Previous methods, beyond architecture decoupling, primarily mitigate task conflicts by adjusting the ratio of understanding and generation data. As concluded in BAGEL~\cite{bagel}, understanding tasks converge faster than generation tasks, leading to a data distribution heavily skewed toward generation in later training stages. In this work, we alleviate task conflicts through the AIA loss and further investigate how this affects the model's sensitivity to data sampling ratios. As shown in \Cref{tab:ablation}, we find that a balanced 1:1 ratio achieves the best performance, contrary to the conventional high-generation, low-understanding distribution. This suggests that with the AIA loss, \textit{understanding and generation tasks not only experience reduced conflict but also exhibit synergistic effects} (\ie, training with both tasks outperforms using generation data alone). We believe this represents a significant step forward in unified model training.

\noindent\textbf{Generality across Model and Data Scales.} While our main experiments focus on the 7B model with 3M data, we further investigate the generality of our approach across different model sizes and data regimes. As shown in \Cref{tab:general}, our method outperforms the baseline in all settings. Notably, even in low-resource scenarios, our approach yields significant gains, demonstrating its robustness and data efficiency. This confirms that AIA is not limited to large-scale training but is a versatile strategy applicable to various computational budgets.

\begin{table*}[t]
    \centering
    \caption{Generality Analysis. We compare the performance of the baseline and our method across different model sizes (1B, 7B) and training data scales (600k, 3M).}
    \vspace{-2mm}
    \resizebox{0.8\linewidth}{!}{
        \begin{tabular}{llcccccccc}
        \toprule[0.15em]
        \multirow{2}{*}{\textbf{Data Size}} & \multirow{2}{*}{\textbf{Method}} & \multicolumn{6}{c}{\textbf{Image Understanding}} & \multicolumn{2}{c}{\textbf{Image Generation}} \\
        \cmidrule(lr){3-8} \cmidrule(lr){9-10}
         & & MMMU & MMBench & MMVP & MMVet & POPE & MME-P & GenEval & DPG  \\
        \midrule[0.1em]
        \multicolumn{10}{l}{\textit{Data-600k}} \\
        \midrule[0.1em]
        \multirow{2}{*}{1B Model} & Janus-Pro & 36.1 & 61.2 & 43.1 & 37.4 & 86.5 & 1448.0 & 0.72 & 82.93 \\
         & \textbf{w AIA} & \textbf{37.2} & \textbf{65.6} & \textbf{43.8} & \textbf{38.5} & \textbf{87.1} & \textbf{1467.4} & \textbf{0.73} & \textbf{83.21} \\
        \cmidrule(lr){2-10}
        \multirow{2}{*}{7B Model} & Janus-Pro & 40.1 & 70.4 & 47.3 & 48.4 & 87.5 & 1587.4 & 0.80 & 84.06 \\
         & \textbf{w AIA} & \textbf{41.1} & \textbf{73.7} & \textbf{48.7} & \textbf{49.0} & \textbf{88.9} & \textbf{1637.8} & \textbf{0.80} & \textbf{84.78} \\
        \midrule[0.1em]
        \multicolumn{10}{l}{\textit{Data-3M}} \\
        \midrule[0.1em]
        \multirow{2}{*}{1B Model} & Janus-Pro & 36.6 & 62.1 & 42.4 & 38.9 & 86.7 & 1424.0 & 0.73 & 83.01 \\
         & \textbf{w AIA} & \textbf{37.7} & \textbf{67.4} & \textbf{43.8} & \textbf{39.4} & \textbf{87.6} & \textbf{1487} & \textbf{0.75} & \textbf{83.67} \\
        \cmidrule(lr){2-10}
        \multirow{2}{*}{7B Model} & Janus-Pro & 40.7 & 71.5 & 47.3 & 49.2 & 88.1 & 1593.1 & 0.80 & 84.19 \\
         & \textbf{w AIA} & \textbf{42.1} & \textbf{75.6} & \textbf{48.0} & \textbf{49.8} & \textbf{89.8} & \textbf{1656.4} & \textbf{0.81} & \textbf{84.49} \\
        
        \bottomrule[0.1em]
        \end{tabular}
    }
    \vspace{-2mm}
    \label{tab:general}
\end{table*}

\subsection{Discussions}

\noindent\textbf{The Task-Specific Attention Patterns Difference in Various Models.} As shown in \cref{fig:cross-attn}, for understanding tasks, models across different series and sizes exhibit nearly identical cross-modal attention patterns due to their shared autoregressive architecture, reflecting the maturity of current understanding architectures. In contrast, generation tasks present diverse patterns due to two distinct architectural paradigms: autoregressive with diffusion head and pure diffusion. Notably, models within the same architectural type display consistent attention patterns. We further observe that generation attention patterns closely follow their underlying attention mechanisms: models with bidirectional attention (FLUX, Qwen-Image, HunyuanImage-3.0) maintain consistent cross-modal interaction intensity across all layers, while those with causal attention (SimpleAR) progressively reduce attention to text as image tokens are generated—completely different from understanding tasks. Interestingly, even with the same attention mechanism, understanding and generation tasks exhibit distinct cross-modal patterns. This difference may hold the key to resolving task conflicts in unified training.

\noindent\textbf{What is the Right Path toward Unified Models?} Through our investigation, we find that regardless of architecture decoupling strategies, generation and understanding tasks consistently exhibit mutually exclusive cross-modal interactions within the same network layers. This aligns with our earlier observation that even under the same autoregressive architecture, understanding and generation display distinct cross-modal patterns when trained independently, suggesting that the task conflicts transcends any architecture design. However, this raises an intriguing question: \textbf{\textit{if models invariably learn task-exclusive interaction patterns during unified training—regardless of architecture choices—could this actually represent the correct behavior for unified models?}} 
Despite the negative correlation between tasks, models can identify the current task through special tokens (\eg, $<$img start$>$), and automatically adjust cross-modal interaction accordingly. With appropriate explicit guidance methods like our AIA, task conflicts may not be an issue to avoid, but rather a natural characteristic to manage.
% Despite the negative correlation between tasks, models can identify the current task through input sequences (image-text or text-image) and special tokens (\eg, $<$img start$>$), and automatically adjust cross-modal interaction accordingly. With appropriate explicit guidance methods like our AIA, task conflicts may not be an issue to avoid, but rather a natural characteristic to manage.

An alternative path toward unification would be unified models with a diffusion head. The core conflict between generation and understanding stems from their divergent output requirements. Understanding tasks demand high-level semantic information, whereas generation necessitates the reconstruction of fine-grained details~\cite{repa}. Even with a unified tokenizer~\cite{tang2025unilip,qu2025tokenflow,vtp,ma2025unitok,lu2025atoken}, this fundamental tension persists. To resolve this, we propose aligning the information density of image and text outputs within the unified model (\ie, this is also mentioned in RecA~\cite{reca}). Specifically, the model generates only abstract conceptual images, which then serve as conditions for the diffusion head to restore fine-grained details. This approach effectively eliminates the conflict between generation and understanding within the unified framework, which is a more realistic way toward real UMM and we take this as future works.

\section{Conclusion}
In this work, we investigated the fundamental challenge of unified multimodal models training: the inherent conflict between image generation and understanding tasks. While existing approaches rely on model decoupling—such as dual image encoders, MOE/MOT architectures, or fixed MLLMs—to mitigate these conflicts, such strategies often sacrifice the interleaved generation capability that defines true unified models. Through systematic analysis of cross-modal attention behaviors, we revealed that model decoupling essentially guides models toward task-specific multimodal interaction patterns, with more aggressive decoupling yielding stronger alignment to single-task behaviors. Building on this insight, we proposed Attention Interaction Alignment (AIA) loss, which explicitly learns task-specific interaction patterns without architecture modifications. We validated AIA on Emu3 during SFT and Janus-Pro during post-training, demonstrating that it not only refines cross-modal attention patterns but also consistently improves both generation and understanding performance across different training stages and architectures. This work represents a significant step toward achieving high-performance unified models while preserving their core capability of seamless cross-modal reasoning.
{
    \small
    \bibliographystyle{unsrt}
    \bibliography{conference}

@String(IJCV  = {Int. J. Comput. Vis.})

@String(CVPR  = {IEEE Conf. Comput. Vis. Pattern Recog.})

@String(ECCV  = {Eur. Conf. Comput. Vis.})

@String(NeurIPS = {Adv. Neural Inform. Process. Syst.})

@String(ICML  = {Int. Conf. Mach. Learn.})

@String(ICLR  = {Int. Conf. Learn. Represent.})

@String(IJCV  = {IJCV})

@String(CVPR  = {CVPR})

@String(ECCV  = {ECCV})

@String(NeurIPS = {NeurIPS})

@String(ICML  = {ICML})

@String(ICLR  = {ICLR})

@article{team2024chameleon,
  title={Chameleon: Mixed-modal early-fusion foundation models},
  author={Team, Chameleon},
  journal={arXiv preprint arXiv:2405.09818},
  year={2024}
}

@inproceedings{qu2025tokenflow,
  title={Tokenflow: Unified image tokenizer for multimodal understanding and generation},
  author={Qu, Liao and Zhang, Huichao and Liu, Yiheng and Wang, Xu and Jiang, Yi and Gao, Yiming and Ye, Hu and Du, Daniel K and Yuan, Zehuan and Wu, Xinglong},
  booktitle={CVPR},
  year={2025}
}

@article{lin2025uniworld,
  title={Uniworld: High-resolution semantic encoders for unified visual understanding and generation},
  author={Lin, Bin and Li, Zongjian and Cheng, Xinhua and Niu, Yuwei and Ye, Yang and He, Xianyi and Yuan, Shenghai and Yu, Wangbo and Wang, Shaodong and Ge, Yunyang and others},
  journal={arXiv preprint arXiv:2506.03147},
  year={2025}
}

@article{Qwen2.5-VL,
  title={Qwen2.5-VL Technical Report},
  author={Bai, Shuai and Chen, Keqin and Liu, Xuejing and Wang, Jialin and Ge, Wenbin and Song, Sibo and Dang, Kai and Wang, Peng and Wang, Shijie and Tang, Jun and Zhong, Humen and Zhu, Yuanzhi and Yang, Mingkun and Li, Zhaohai and Wan, Jianqiang and Wang, Pengfei and Ding, Wei and Fu, Zheren and Xu, Yiheng and Ye, Jiabo and Zhang, Xi and Xie, Tianbao and Cheng, Zesen and Zhang, Hang and Yang, Zhibo and Xu, Haiyang and Lin, Junyang},
  journal={arXiv preprint arXiv:2502.13923},
  year={2025}
}

@article{cao2025hunyuanimage,
  title={Hunyuanimage 3.0 technical report},
  author={Cao, Siyu and Chen, Hangting and Chen, Peng and Cheng, Yiji and Cui, Yutao and Deng, Xinchi and Dong, Ying and Gong, Kipper and Gu, Tianpeng and Gu, Xiusen and others},
  journal={arXiv preprint arXiv:2509.23951},
  year={2025}
}

@article{qwenimage,
  title={Qwen-image technical report},
  author={Wu, Chenfei and Li, Jiahao and Zhou, Jingren and Lin, Junyang and Gao, Kaiyuan and Yan, Kun and Yin, Sheng-ming and Bai, Shuai and Xu, Xiao and Chen, Yilei and others},
  journal={arXiv preprint arXiv:2508.02324},
  year={2025}
}

@article{metaquery,
  title={Transfer between modalities with metaqueries},
  author={Pan, Xichen and Shukla, Satya Narayan and Singh, Aashu and Zhao, Zhuokai and Mishra, Shlok Kumar and Wang, Jialiang and Xu, Zhiyang and Chen, Jiuhai and Li, Kunpeng and Juefei-Xu, Felix and Hou, Ji and Xie, Saining},
  journal={arXiv preprint arXiv:2504.06256},
  year={2025}
}

@article{podell2023sdxl,
  title={Sdxl: Improving latent diffusion models for high-resolution image synthesis},
  author={Podell, Dustin and English, Zion and Lacey, Kyle and Blattmann, Andreas and Dockhorn, Tim and M{\"u}ller, Jonas and Penna, Joe and Rombach, Robin},
  journal={arXiv preprint arXiv:2307.01952},
  year={2023}
}

@inproceedings{sd3,
  title={Scaling rectified flow transformers for high-resolution image synthesis},
  author={Esser, Patrick and Kulal, Sumith and Blattmann, Andreas and Entezari, Rahim and M{\"u}ller, Jonas and Saini, Harry and Levi, Yam and Lorenz, Dominik and Sauer, Axel and Boesel, Frederic and others},
  booktitle={ICML},
  year={2024}
}

@inproceedings{liu2025shotbench,
  title={ShotBench: Expert-Level Cinematic Understanding in Vision-Language Models},
  author={Liu, Hongbo and He, Jingwen and Jin, Yi and Zheng, Dian and Dong, Yuhao and Zhang, Fan and Huang, Ziqi and He, Yinan and Li, Yangguang and Chen, Weichao and others},
  booktitle={NeurIPS},
  year={2025}
}

@article{unimmmu,
  title={Uni-MMMU: A Massive Multi-discipline Multimodal Unified Benchmark},
  author={Zou, Kai and Huang, Ziqi and Dong, Yuhao and Tian, Shulin and Zheng, Dian and Liu, Hongbo and He, Jingwen and Liu, Bin and Qiao, Yu and Liu, Ziwei},
  journal={arXiv preprint arXiv:2510.13759},
  year={2025}
}

@misc{flux2024,
    author={Black Forest Labs},
    title={FLUX},
    year={2024},
    howpublished={\url{https://github.com/black-forest-labs/flux}},
}

@article{tang2025unilip,
  title={Unilip: Adapting clip for unified multimodal understanding, generation and editing},
  author={Tang, Hao and Xie, Chenwei and Bao, Xiaoyi and Weng, Tingyu and Li, Pandeng and Zheng, Yun and Wang, Liwei},
  journal={arXiv preprint arXiv:2507.23278},
  year={2025}
}

@article{vae,
  title={Auto-encoding variational bayes},
  author={Kingma, Diederik P and Welling, Max},
  journal={arXiv preprint arXiv:1312.6114},
  year={2013}
}

@article{wu2024vila,
  title={Vila-u: a unified foundation model integrating visual understanding and generation},
  author={Wu, Yecheng and Zhang, Zhuoyang and Chen, Junyu and Tang, Haotian and Li, Dacheng and Fang, Yunhao and Zhu, Ligeng and Xie, Enze and Yin, Hongxu and Yi, Li and others},
  journal={arXiv preprint arXiv:2409.04429},
  year={2024}
}

@inproceedings{han2025infinity,
  title={Infinity: Scaling bitwise autoregressive modeling for high-resolution image synthesis},
  author={Han, Jian and Liu, Jinlai and Jiang, Yi and Yan, Bin and Zhang, Yuqi and Yuan, Zehuan and Peng, Bingyue and Liu, Xiaobing},
  booktitle={CVPR},
  year={2025}
}

@article{wu2024janus,
  title={Janus: Decoupling visual encoding for unified multimodal understanding and generation},
  author={Wu, Chengyue and Chen, Xiaokang and Wu, Zhiyu and Ma, Yiyang and Liu, Xingchao and Pan, Zizheng and Liu, Wen and Xie, Zhenda and Yu, Xingkai and Ruan, Chong and others},
  journal={arXiv preprint arXiv:2410.13848},
  year={2024}
}

@article{liu2024deepseek,
  title={Deepseek-v3 technical report},
  author={Liu, Aixin and Feng, Bei and Xue, Bing and Wang, Bingxuan and Wu, Bochao and Lu, Chengda and Zhao, Chenggang and Deng, Chengqi and Zhang, Chenyu and Ruan, Chong and others},
  journal={arXiv preprint arXiv:2412.19437},
  year={2024}
}

@article{fu2023mme,
  title={MME: A Comprehensive Evaluation Benchmark for Multimodal Large Language Models},
  author={Fu, Chaoyou and Chen, Peixian and Shen, Yunhang and Qin, Yulei and Zhang, Mengdan and Lin, Xu and Yang, Jinrui and Zheng, Xiawu and Li, Ke and Sun, Xing and others},
  journal={arXiv preprint arXiv:2306.13394},
  year={2023}
}

@inproceedings{ghosh2023geneval,
  title={Geneval: An object-focused framework for evaluating text-to-image alignment},
  author={Ghosh, Dhruba and Hajishirzi, Hannaneh and Schmidt, Ludwig},
  booktitle={NeurIPS},
  year={2023}
}

@inproceedings{mmvp,
  title={Eyes wide shut? exploring the visual shortcomings of multimodal llms},
  author={Tong, Shengbang and Liu, Zhuang and Zhai, Yuexiang and Ma, Yi and LeCun, Yann and Xie, Saining},
  booktitle={CVPR},
  year={2024}
}

@inproceedings{yue2024mmmu,
  title={Mmmu: A massive multi-discipline multimodal understanding and reasoning benchmark for expert agi},
  author={Yue, Xiang and Ni, Yuansheng and Zhang, Kai and Zheng, Tianyu and Liu, Ruoqi and Zhang, Ge and Stevens, Samuel and Jiang, Dongfu and Ren, Weiming and Sun, Yuxuan and others},
  booktitle={CVPR},
  year={2024}
}

@article{zhu2025internvl3,
  title={Internvl3: Exploring advanced training and test-time recipes for open-source multimodal models},
  author={Zhu, Jinguo and Wang, Weiyun and Chen, Zhe and Liu, Zhaoyang and Ye, Shenglong and Gu, Lixin and Tian, Hao and Duan, Yuchen and Su, Weijie and Shao, Jie and others},
  journal={arXiv preprint arXiv:2504.10479},
  year={2025}
}

@inproceedings{chen2024internvl,
  title={Internvl: Scaling up vision foundation models and aligning for generic visual-linguistic tasks},
  author={Chen, Zhe and Wu, Jiannan and Wang, Wenhai and Su, Weijie and Chen, Guo and Xing, Sen and Zhong, Muyan and Zhang, Qinglong and Zhu, Xizhou and Lu, Lewei and others},
  booktitle={CVPR},
  year={2024}
}

@article{reca,
  title={Reconstruction alignment improves unified multimodal models},
  author={Xie, Ji and Darrell, Trevor and Zettlemoyer, Luke and Wang, XuDong},
  journal={arXiv preprint arXiv:2509.07295},
  year={2025}
}

@article{wang2025sparsemm,
  title={SparseMM: Head Sparsity Emerges from Visual Concept Responses in MLLMs},
  author={Wang, Jiahui and Liu, Zuyan and Rao, Yongming and Lu, Jiwen},
  journal={arXiv preprint arXiv:2506.05344},
  year={2025}
}

@article{zhang2024sparsevlm,
  title={Sparsevlm: Visual token sparsification for efficient vision-language model inference},
  author={Zhang, Yuan and Fan, Chun-Kai and Ma, Junpeng and Zheng, Wenzhao and Huang, Tao and Cheng, Kuan and Gudovskiy, Denis and Okuno, Tomoyuki and Nakata, Yohei and Keutzer, Kurt and others},
  journal={arXiv preprint arXiv:2410.04417},
  year={2024}
}

@article{pope,
  title={Evaluating object hallucination in large vision-language models},
  author={Li, Yifan and Du, Yifan and Zhou, Kun and Wang, Jinpeng and Zhao, Wayne Xin and Wen, Ji-Rong},
  journal={arXiv preprint arXiv:2305.10355},
  year={2023}
}

@article{dpg,
  title={Ella: Equip diffusion models with llm for enhanced semantic alignment},
  author={Hu, Xiwei and Wang, Rui and Fang, Yixiao and Fu, Bin and Cheng, Pei and Yu, Gang},
  journal={arXiv preprint arXiv:2403.05135},
  year={2024}
}

@article{mmvet,
  title={Mm-vet: Evaluating large multimodal models for integrated capabilities},
  author={Yu, Weihao and Yang, Zhengyuan and Li, Linjie and Wang, Jianfeng and Lin, Kevin and Liu, Zicheng and Wang, Xinchao and Wang, Lijuan},
  journal={arXiv preprint arXiv:2308.02490},
  year={2023}
}

@inproceedings{liu2024mmbench,
  title={Mmbench: Is your multi-modal model an all-around player?},
  author={Liu, Yuan and Duan, Haodong and Zhang, Yuanhan and Li, Bo and Zhang, Songyang and Zhao, Wangbo and Yuan, Yike and Wang, Jiaqi and He, Conghui and Liu, Ziwei and others},
  booktitle={ECCV},
  year={2024},
}

@article{chen2025sharegpt,
  title={ShareGPT-4o-Image: Aligning Multimodal Models with GPT-4o-Level Image Generation},
  author={Chen, Junying and Cai, Zhenyang and Chen, Pengcheng and Chen, Shunian and Ji, Ke and Wang, Xidong and Yang, Yunjin and Wang, Benyou},
  journal={arXiv preprint arXiv:2506.18095},
  year={2025}
}

@article{zheng2024open,
  title={Open-sora: Democratizing efficient video production for all},
  author={Zheng, Zangwei and Peng, Xiangyu and Yang, Tianji and Shen, Chenhui and Li, Shenggui and Liu, Hongxin and Zhou, Yukun and Li, Tianyi and You, Yang},
  journal={arXiv preprint arXiv:2412.20404},
  year={2024}
}

@article{guo2024mammoth,
  title={Mammoth-vl: Eliciting multimodal reasoning with instruction tuning at scale},
  author={Guo, Jarvis and Zheng, Tuney and Bai, Yuelin and Li, Bo and Wang, Yubo and Zhu, King and Li, Yizhi and Neubig, Graham and Chen, Wenhu and Yue, Xiang},
  journal={arXiv preprint arXiv:2412.05237},
  year={2024}
}

@inproceedings{ddim,
  title={Denoising diffusion implicit models},
  author={Song, Jiaming and Meng, Chenlin and Ermon, Stefano},
  booktitle={ICLR},
  year={2020}
}

@inproceedings{ddpm,
  title={Denoising diffusion probabilistic models},
  author={Ho, Jonathan and Jain, Ajay and Abbeel, Pieter},
  booktitle={NeurIPS},
  year={2020}
}

@article{yang2025qwen3,
  title={Qwen3 technical report},
  author={Yang, An and Li, Anfeng and Yang, Baosong and Zhang, Beichen and Hui, Binyuan and Zheng, Bo and Yu, Bowen and Gao, Chang and Huang, Chengen and Lv, Chenxu and others},
  journal={arXiv preprint arXiv:2505.09388},
  year={2025}
}

@article{team2024qwen2,
  title={Qwen2 technical report},
  author={Team, Qwen and others},
  journal={arXiv preprint arXiv:2407.10671},
  year={2024}
}

@article{li2024llava,
  title={Llava-onevision: Easy visual task transfer},
  author={Li, Bo and Zhang, Yuanhan and Guo, Dong and Zhang, Renrui and Li, Feng and Zhang, Hao and Zhang, Kaichen and Zhang, Peiyuan and Li, Yanwei and Liu, Ziwei and others},
  journal={arXiv preprint arXiv:2408.03326},
  year={2024}
}

@misc{ma2024janusflow,
      title={JanusFlow: Harmonizing Autoregression and Rectified Flow for Unified Multimodal Understanding and Generation}, 
      author={Yiyang Ma and Xingchao Liu and Xiaokang Chen and Wen Liu and Chengyue Wu and Zhiyu Wu and Zizheng Pan and Zhenda Xie and Haowei Zhang and Xingkai yu and Liang Zhao and Yisong Wang and Jiaying Liu and Chong Ruan},
      journal={arXiv preprint arXiv:2411.07975},
      year={2024}
}

@inproceedings{clip,
  title={Learning transferable visual models from natural language supervision},
  author={Radford, Alec and Kim, Jong Wook and Hallacy, Chris and Ramesh, Aditya and Goh, Gabriel and Agarwal, Sandhini and Sastry, Girish and Askell, Amanda and Mishkin, Pamela and Clark, Jack and others},
  booktitle={ICML},
  year={2021},
}

@article{wang2025simplear,
  title={Simplear: Pushing the frontier of autoregressive visual generation through pretraining, sft, and rl},
  author={Wang, Junke and Tian, Zhi and Wang, Xun and Zhang, Xinyu and Huang, Weilin and Wu, Zuxuan and Jiang, Yu-Gang},
  journal={arXiv preprint arXiv:2504.11455},
  year={2025}
}

@inproceedings{vqvae,
  title={Neural discrete representation learning},
  author={Van Den Oord, Aaron and Vinyals, Oriol and others},
  booktitle={NeurIPS},
  year={2017}
}

@inproceedings{vqgan,
  title={Taming transformers for high-resolution image synthesis},
  author={Esser, Patrick and Rombach, Robin and Ommer, Bjorn},
  booktitle={CVPR},
  year={2021}
}

@article{dubey2024llama,
  title={The llama 3 herd of models},
  author={Dubey, Abhimanyu and Jauhri, Abhinav and Pandey, Abhinav and Kadian, Abhishek and Al-Dahle, Ahmad and Letman, Aiesha and Mathur, Akhil and Schelten, Alan and Yang, Amy and Fan, Angela and others},
  journal={arXiv e-prints},
  year={2024}
}

@article{wu2024liquid,
  title={Liquid: Language models are scalable and unified multi-modal generators},
  author={Wu, Junfeng and Jiang, Yi and Ma, Chuofan and Liu, Yuliang and Zhao, Hengshuang and Yuan, Zehuan and Bai, Song and Bai, Xiang},
  journal={IJCV},
  year={2024}
}

@article{chen2025blip3,
  title={Blip3-o: A family of fully open unified multimodal models-architecture, training and dataset},
  author={Chen, Jiuhai and Xu, Zhiyang and Pan, Xichen and Hu, Yushi and Qin, Can and Goldstein, Tom and Huang, Lifu and Zhou, Tianyi and Xie, Saining and Savarese, Silvio and others},
  journal={arXiv preprint arXiv:2505.09568},
  year={2025}
}

@inproceedings{showo,
  title={Show-o: One single transformer to unify multimodal understanding and generation},
  author={Xie, Jinheng and Mao, Weijia and Bai, Zechen and Zhang, David Junhao and Wang, Weihao and Lin, Kevin Qinghong and Gu, Yuchao and Chen, Zhijie and Yang, Zhenheng and Shou, Mike Zheng},
  booktitle={ICLR},
  year={2025}
}

@inproceedings{repa,
  title={Representation alignment for generation: Training diffusion transformers is easier than you think},
  author={Yu, Sihyun and Kwak, Sangkyung and Jang, Huiwon and Jeong, Jongheon and Huang, Jonathan and Shin, Jinwoo and Xie, Saining},
  booktitle={ICLR},
  year={2025}
}

@inproceedings{ma2025unitok,
  title={Unitok: A unified tokenizer for visual generation and understanding},
  author={Ma, Chuofan and Jiang, Yi and Wu, Junfeng and Yang, Jihan and Yu, Xin and Yuan, Zehuan and Peng, Bingyue and Qi, Xiaojuan},
  booktitle={NeurIPS},
  year={2025}
}

@article{zhou2024transfusion,
  title={Transfusion: Predict the next token and diffuse images with one multi-modal model},
  author={Zhou, Chunting and Yu, Lili and Babu, Arun and Tirumala, Kushal and Yasunaga, Michihiro and Shamis, Leonid and Kahn, Jacob and Ma, Xuezhe and Zettlemoyer, Luke and Levy, Omer},
  journal={arXiv preprint arXiv:2408.11039},
  year={2024}
}

@article{hinton2015distilling,
  title={Distilling the knowledge in a neural network},
  author={Hinton, Geoffrey and Vinyals, Oriol and Dean, Jeff},
  journal={arXiv preprint arXiv:1503.02531},
  year={2015}
}

@article{bai2025qwen3vl,
  title={Qwen3-vl technical report},
  author={Bai, Shuai and Cai, Yuxuan and Chen, Ruizhe and Chen, Keqin and Chen, Xionghui and Cheng, Zesen and Deng, Lianghao and Ding, Wei and Gao, Chang and Ge, Chunjiang and others},
  journal={arXiv preprint arXiv:2511.21631},
  year={2025}
}

@article{fitnet,
  title={Fitnets: Hints for thin deep nets},
  author={Romero, Adriana and Ballas, Nicolas and Kahou, Samira Ebrahimi and Chassang, Antoine and Gatta, Carlo and Bengio, Yoshua},
  journal={arXiv preprint arXiv:1412.6550},
  year={2014}
}

@article{longcatimage,
  title={Longcat-image technical report},
  author={Team, Meituan LongCat and Ma, Hanghang and Tan, Haoxian and Huang, Jiale and Wu, Junqiang and He, Jun-Yan and Gao, Lishuai and Xiao, Songlin and Wei, Xiaoming and Ma, Xiaoqi and others},
  journal={arXiv preprint arXiv:2512.07584},
  year={2025}
}

@article{vtp,
  title={Towards Scalable Pre-training of Visual Tokenizers for Generation},
  author={Yao, Jingfeng and Song, Yuda and Zhou, Yucong and Wang, Xinggang},
  journal={arXiv preprint arXiv:2512.13687},
  year={2025}
}

@article{lu2025atoken,
  title={Atoken: A unified tokenizer for vision},
  author={Lu, Jiasen and Song, Liangchen and Xu, Mingze and Ahn, Byeongjoo and Wang, Yanjun and Chen, Chen and Dehghan, Afshin and Yang, Yinfei},
  journal={arXiv preprint arXiv:2509.14476},
  year={2025}
}

@article{wang2024emu3,
  title={Emu3: Next-token prediction is all you need},
  author={Wang, Xinlong and Zhang, Xiaosong and Luo, Zhengxiong and Sun, Quan and Cui, Yufeng and Wang, Jinsheng and Zhang, Fan and Wang, Yueze and Li, Zhen and Yu, Qiying and others},
  journal={arXiv preprint arXiv:2409.18869},
  year={2024}
}

@inproceedings{xie2025showo2,
  title={Show-o2: Improved Native Unified Multimodal Models},
  author={Xie, Jinheng and Yang, Zhenheng and Shou, Mike Zheng},
  booktitle={NeurIPS},
  year={2025}
}

@article{li2025onecat,
  title={Onecat: Decoder-only auto-regressive model for unified understanding and generation},
  author={Li, Han and Peng, Xinyu and Wang, Yaoming and Peng, Zelin and Chen, Xin and Weng, Rongxiang and Wang, Jingang and Cai, Xunliang and Dai, Wenrui and Xiong, Hongkai},
  journal={arXiv preprint arXiv:2509.03498},
  year={2025}
}

@article{januspro,
  title={Janus-pro: Unified multimodal understanding and generation with data and model scaling},
  author={Chen, Xiaokang and Wu, Zhiyu and Liu, Xingchao and Pan, Zizheng and Liu, Wen and Xie, Zhenda and Yu, Xingkai and Ruan, Chong},
  journal={arXiv preprint arXiv:2501.17811},
  year={2025}
}

@article{yang2025survey,
  title={A Survey of Unified Multimodal Understanding and Generation: Advances and Challenges},
  author={Yang, Yan and Tian, Haochen and Shi, Yang and Xie, Wulin and Zhang, Yi-Fan and Dong, Yuhao and Hu, Yibo and Wang, Liang and He, Ran and Shan, Caifeng and others},
  journal={Authorea Preprints},
  year={2025},
}

@article{geng2025x,
  title={X-omni: Reinforcement learning makes discrete autoregressive image generative models great again},
  author={Geng, Zigang and Wang, Yibing and Ma, Yeyao and Li, Chen and Rao, Yongming and Gu, Shuyang and Zhong, Zhao and Lu, Qinglin and Hu, Han and Zhang, Xiaosong and others},
  journal={arXiv preprint arXiv:2507.22058},
  year={2025}
}

@article{shi2025realunify,
  title={Realunify: Do unified models truly benefit from unification? a comprehensive benchmark},
  author={Shi, Yang and Dong, Yuhao and Ding, Yue and Wang, Yuran and Zhu, Xuanyu and Zhou, Sheng and Liu, Wenting and Tian, Haochen and Wang, Rundong and Wang, Huanqian and others},
  journal={arXiv preprint arXiv:2509.24897},
  year={2025}
}

@article{zimage,
  title={Z-image: An efficient image generation foundation model with single-stream diffusion transformer},
  author={Cai, Huanqia and Cao, Sihan and Du, Ruoyi and Gao, Peng and Hoi, Steven and Hou, Zhaohui and Huang, Shijie and Jiang, Dengyang and Jin, Xin and Li, Liangchen and others},
  journal={arXiv preprint arXiv:2511.22699},
  year={2025}
}

@article{bagel,
  title={Emerging properties in unified multimodal pretraining},
  author={Deng, Chaorui and Zhu, Deyao and Li, Kunchang and Gou, Chenhui and Li, Feng and Wang, Zeyu and Zhong, Shu and Yu, Weihao and Nie, Xiaonan and Song, Ziang and others},
  journal={arXiv preprint arXiv:2505.14683},
  year={2025}
}

@article{wu2025omnigen2,
  title={OmniGen2: Exploration to Advanced Multimodal Generation},
  author={Wu, Chenyuan and Zheng, Pengfei and Yan, Ruiran and Xiao, Shitao and Luo, Xin and Wang, Yueze and Li, Wanli and Jiang, Xiyan and Liu, Yexin and Zhou, Junjie and others},
  journal={arXiv preprint arXiv:2506.18871},
  year={2025}
}
}

\end{document}